\documentclass{article}
\usepackage{amsmath, amssymb}
\usepackage{listings}
\usepackage{graphicx}
\usepackage{color}
\begin{document}
\lstset{basicstyle=\footnotesize,breaklines=true}
\newcommand{\fakeincludecodefigure}[2]{\begin{figure}\begin{center}Eventually the code file \lstinline|#1| will be listed here.\end{center}\caption{#2}\label{fig:#1}\end{figure}}
\newcommand{\includecodefigure}[2]{\begin{figure}\begin{center}\lstinputlisting[basicstyle=\ttfamily, columns=fullflexible, keepspaces=true]{codeinclude/#1.txt}\end{center}\caption{#2}\label{fig:#1}\end{figure}}
\newcommand{\includegraphicsfigure}[3]{\begin{figure}\begin{center}\includegraphics[#3]{include/#1.pdf}\end{center}\caption{#2}\label{fig:#1}\end{figure}}
\newcommand{\fakeincludegraphicsfigure}[3]{\begin{figure}\begin{center}Eventually the graphic \lstinline|#1| will be displayed here.\end{center}\caption{#2}\label{fig:#1}\end{figure}}
\newcommand{\footremember}[2]{
    \footnote{#2}
    \newcounter{#1}
    \setcounter{#1}{\value{footnote}}
}
\newcommand{\footrecall}[1]{%
    \footnotemark[\value{#1}]%
}
\lstset{language=Python}
\title{Solving differential equations with unknown constitutive relations as recurrent neural networks}
\author{Tobias Hagge\footremember{PNNL}{Pacific Northwest National Laboratory} \and Panos Stinis\footrecall{PNNL} \and Enoch Yeung\footrecall{PNNL} \and Alexandre M. Tartakovsky\footrecall{PNNL}}
\maketitle

\begin{abstract}
  We solve a system of ordinary differential equations with an unknown functional form of a sink (reaction rate) term. We assume that the measurements (time series) of state variables are partially available, and we use recurrent neural network to ``learn'' the reaction rate from this data. This is achieved by including a  discretized ordinary differential equations as part of a recurrent neural network training problem. We extend TensorFlow's recurrent neural network architecture to create a simple but scalable and effective solver for the unknown functions, and apply it to a fedbatch bioreactor simulation problem. Use of techniques from recent deep learning literature enables training of functions with behavior manifesting over thousands of time steps. Our networks are structurally similar to recurrent neural networks, but differences in design and function require modifications to the conventional wisdom about training such networks.
  
\end{abstract}

\section{Introduction}\label{sec:intro}
Neural networks have an extensive history as tools for numerical solution of differential equations, with recurrent neural networks playing a role from the start. An early example was \cite{lee-kang-1990}, which described how to construct, given a set of solvable difference equations, a Hopfield network, the minimal energy states of which are time series satisfying those differential equations. Scalability was advertised as a feature of these networks; the authors argued that their algorithm was highly parallel and relied only on simple operations.

We are interested in scalable solutions to the problem of training neural networks to estimate unknown functions in ordinary differential equations (ODEs). An early application of neural networks for this purpose is found in the 1992 work of Psichogios and Ungar \cite{psichogios-ungar}, in which a fedbatch reactor model predicts total volume, substrate volume, and reactant volume over time. In this work, an  ODE model is used to describe mass conservation with an unknown kinetic term $\mu$, which is represented as a two-layer feed-forward neural network (NN) with sigmoidal activations. Psichogios and Ungar demonstrated that their model could learn $\mu$ correctly even when the reactant volume is unmeasurable, but at the cost of solving additional sensitivity ODEs. Also, they gave their model an advantage by providing a training corpus over which the domain of $\mu$ was well-covered by the first few time steps, leaving open the question of whether long time-series behavior can be used to train an unknown function in a process of known functional form.

In recent years, neural networks have experienced a resurgence in machine learning due to processing power advances and the development of so-called 'deep learning' techniques for designing, initializing, and training many-layer networks. These techniques help overcome a network training issue known as the {\em exploding and vanishing gradients problem}, which causes training rates for many-layer neural networks to become infeasibly slow. Deep networks in machine learning are multiply motivated; in image processing there is a need for networks that can learn multiple hierarchies of abstraction in order to recognize image features; in machine translation applications, meaning is embodied in combinations of words which are distant from one another.

In the case of process modeling, there are at least two motivations for ``deep'' hidden-state process models. The first motivation is empirical: in some applications, some variables cannot be measured quickly and/or inexpensively thus are unsuitable as model input. The second motivation is statistical: when one has access to every observable at every time-step, it may be tempting to treat each pair of consecutive states as a training sample. However, if the function is nonlinear, unbiased noise in the data can bias the training of the function. Such bias can be greatly reduced if the system is required to maintain the correct behavior over many time-steps.

In this work, we re-examine the problem of Psichogios and Ungar in light of recent advances in deep learning research. The authors of that work considered the network training problem as an application of the sensitivity equations, and it was necessary to integrate the neural network training software with a differential equation solver in order to backpropagate the difference of predicted and observed vales. We recast the problem as a recurrent neural network training problem. This formulation eliminates the need for solving sensitivity equations as a part of the backpropagation procedure, and with some care, techniques from deep learning can be applied to train functions depending on long term time-series behavior.

As processor clock speed increases have slowed, performance improvements have increasingly relied on parallelization, with HPC systems possessing many CPU and/or GPU cores. The domains of process modeling and machine learning share the need for scalable software capable of efficiently using parallel resources in heterogeneous environments without placing undue burden on the researcher. Our implementation extends the recurrent neural network architecture of Google's TensorFlow software package \cite{tensorflow2015-whitepaper} to solve difference equations (resulting from discretization of the governing ODEs) containing unknown functions. It benefits from the scalability and maturity of that package. We demonstrate the ability of our solver to train unknown functions which require hundreds or thousands of time-steps to exercise their behavior, and highlight some of its useful properties. Our networks are formally similar to unrolled recurrent neural networks, but different in design: while layer architectures such as LSTM and GRU are structured to retain memory and to overcome the vanishing and exploding gradients problem, our layers enforce dynamical constraints. As a result, some of the conventional wisdom about recurrent neural networks must be modified when our case is considered.

The structure of the remaining sections is as follows. In Section~\ref{sec:tensorflow} we describe TensorFlow's general architecture and its facilities for recurrent neural networks. Readers familiar with TensorFlow may choose to skip this section. In Section~\ref{sec:solver} we describe our solver. In Section~\ref{sec:problem} we describe the structure and implementation of the fedbatch reactor problem we address. In Section~\ref{sec:results} we discuss simulation results demonstrating our solver's capabilities and properties. In Section~\ref{sec:comparison} we compare training issues in the proposed network with those for other recurrent neural networks, providing some background along with references. 

\section{TensorFlow}\label{sec:tensorflow}
\subsection{Basics}
TensorFlow\cite{tensorflow2015-whitepaper} is an open source software library for scalable, vectorized numerical computation. TensorFlow uses a graph model to describe computations, and a backend to deploy graph-modeled computations to CPUs and/or GPUs. CPU/GPU allocation and scheduling are handled automatically unless the programmer intervenes. TensorFlow uses an accelerated linear algebra library called XLA on the back end to perform computations efficiently and scalably. At present, TensorFlow does not automatically scale computations beyond a single memory space, although such scaling can be implemented in TensorFlow if additional code is written. Third party TensorFlow extensions such as MATEX-TensorFlow \cite{matex-tensorflow} enable cluster-level parallelism without the need for extra code using a data-parallel approach.

As of this writing, there are TensorFlow APIs for the Python, C++, Java, and Go programming languages. The Python API is currently the most feature-complete and mature, and is the implementation used and discussed in this work. For best performance it is desirable to embed the entirety of a computation within a TensorFlow graph, however it is often more convenient to write Python code which invokes TensorFlow graph operations as needed, trading some performance for convenience.

Computation in TensorFlow is a two stage process. In the first stage, one constructs a computational {\em graph}. The nodes of the graph represent operations (in TensorFlow parlance: {\em ops}). Directed edges represent data, which is input to or output from the operations. The inputs to the computation as a whole are represented by nodes known as {\em placeholders}. TensorFlow graphs are append-only; there is no mechanism for deleting nodes or edges. The typical mode of operation is to construct the graph in its entirety prior to performing any computations.

TensorFlow graphs do not instantiate a computation in hardware or maintain state. In the second stage of computation, one constructs and initializes a run-time instance known as a {\em session} to perform these functions. After initialization, one performs computations by invoking the session's \lstinline|run| method, specifying the nodes to be evaluated along with values for the necessary placeholders. The back end then evaluates that portion of the instantiated graph necessary to evaluate the specified nodes and returns the results of the evaluations.

The fundamental data type in TensorFlow is a multidimensional array, or {\em tensor}. Operations take tensors as input and return tensors as output. In line with the two-step computational process, tensors are not arrays themselves, but descriptions of arrays which take values in the context of a session. The data types of the returned values are implementation-dependent; in Python they are Numpy arrays. The number of dimensions in a tensor is fixed, but TensorFlow allows dimensions of unknown size in order to better support batch processing and variable-length sequential input.

Tensors are represented in Python by variables of special type. Standard python operators have been overloaded for these data-types to serve as graph construction operators. This is confusing, initially, but convenient; to construct a graph for most functions one writes code syntactically identical to Python code which would execute the function. For example, if \lstinline|a|, \lstinline|b|, and \lstinline|c| are tensors, the Python invocation \lstinline|a = b + c| constructs a \lstinline|tf.add| op node (represented in python by the variable \lstinline|a|)  and adds directed edges from \lstinline|b| and \lstinline|c| into it, then feeds the result to a \lstinline|tf.assign| operator which creates the graph op assigning the result to \lstinline|a|, which must be of type \lstinline|tf.Variable|. This code can later be executed by instantiating and initializing a session, and calling its \lstinline|run| method.

\subsection{Recurrent neural networks in TensorFlow}\label{sec:rnn}
There are multiple ways to implement a recurrent neural network in TensorFlow. One method is to initialize an object of class \lstinline|dynamic_rnn|, passing an object which subclasses \lstinline|RNNCell| to define the behavior of a recurrent layer. Subclasses of \lstinline|RNNCell| implement standard recurrent neural network layers such as LSTM and GRU layers, but the abstraction also supports implementation of other discrete dynamical systems.

Layers in a recurrent neural network have self-connections, which makes backpropagation of error an infinite process. To prevent this, recurrent neural networks are typically implemented via finite approximations. The job of \lstinline|dynamic_rnn| is to assemble a graph which implements this approximation as a finite-time discrete dynamical system. In particular, an invocation of \lstinline|dynamic_rnn| is passed an instance of \lstinline|RNNCell| and a positive integer $n$ and constructs a graph which, at time step $t \in (0 \ldots n-1)$, takes as input a state tensor $s_{t-1}$ from time $t-1$ and an input tensor $i_t$ and outputs a state tensor $s_t$ and an output tensor $o_t$. The \lstinline|RNNCell| instance is responsible for constructing the subgraph defining $s_t$ and $o_t$ from $s_{t-1}$ and $i_t$. The entire graph takes as input an initial-state tensor $s_0$ and an input tensor $i = [i_0, \ldots, i_n]$ and produces as output a final-state tensor $s_n$ and an output tensor $o = [o_0, \ldots, o_n]$.

Networks are evaluated in batch mode using vectorized operations; these tensors are two-dimensional, the size of the first dimension being the number of elements in a batch.

Importantly for our purposes, all trainable parameters, activation functions, etc. are defined in \lstinline|RNNCell|, not in \lstinline|dynamic_rnn|. Thus, \lstinline|dynamic_rnn| implements a batch-mode discrete dynamical system, with neural network architectural elements specified in \lstinline|RNNCell|.

\subsection{Parameter optimization in TensorFlow}

TensorFlow was originally developed for machine learning research, and has a large library of codes developed for this purpose. TensorFlow's ops have built-in support for (reverse) automatic differentiation. As a result, any TensorFlow graph defined with trainable parameters can be trained using standard backpropagation-based optimization techniques, without the need to write any code. All one needs to do is to define a loss function and specify the optimization method and method-specific parameters to be used.

\section{ODE solver implementation}\label{sec:solver}
Our solver implements a subclass \lstinline|EulerCell| of the TensorFlow python class \lstinline|RNNCell|. An instance of this class is passed to TensorFlow's \lstinline|dynamic_rnn| method to construct a graph which performs Euler integration of an ODE. The work is performed by \lstinline|EulerCell|'s \lstinline|__call__| method, the implementation of which is shown in Figure~\ref{fig:euler-cell}.

\includecodefigure{euler-cell}{The \lstinline|__call__| method of the class \lstinline|EulerCell| implements one time step of an Euler integration of a differential equation.}

An \lstinline|RNNCell| \lstinline|__call__| method returns an output as well as a propagated internal state. In our case, the internal state is physical state of the system and is set equal to the output.

Implementation of more sophisticated integration methods, such as Runge Kutta methods, is possible and relatively straightforward.

\section{Fedbatch bioreactor model}\label{sec:problem}

We test our approach on a system of ODEs describing the bioreactor \cite{psichogios-ungar}:
The dynamics of the system is described by the ODEs:
\begin{equation}\label{eq_X}
\frac{\partial X(t)}{\partial t} = \mu(t) X(t) - \frac{F(t) X(t)}{V(t)},
\end{equation}
\begin{equation}\label{eq_S}
\frac{\partial S(t)}{\partial t} = -k_1 \mu(t) X(t) +  \frac{F(t) (S_{in}(t) - S(t))}{V(t)},
\end{equation}
\begin{equation}\label{eq_V}
\frac{\partial V(t)}{\partial t} = F(t),
\end{equation}
where $S$ is the substrate concentration, $X$ is the reactant concentration,  $V$ is the total material volume. The material is fed in with a known rate $F(t)$ and substrate concentration $S_{in}(t)$.
The reaction rate $\mu$ is assumed to be an unknown function of both state variables $X$ and $S$. Our goal is to find a good approximation of $\mu$ based on observations or partial observations of $S$ and $X$ over a range of time steps. Without loss of generality, we assume that the feed rate $F$ is a constant. 

Following \cite{psichogios-ungar}, for the ground truth we use the Haldane model:
\begin{equation}
\mu_g(X, S) = \frac{S \mu^*}{ S +  K_m + \frac{S^2}{K_i}}.
\end{equation}
Here $\mu^*$, $K_m$, and $K_i$ are constants. Note that $\mu_g$ does not depend on $X$.

If the domain of $\mu$ is well-covered by the initial time-steps of a long-time-series data set, it is debatable, and situational, whether there is any point in training the series in their entirety, rather than just over the initial time-steps. Accordingly, we have chosen values for $k_1$, $\mu^*$, $K_m$, and $K_i$, along with distributions for the initial values $S_0$, $X_0$, $V_0$ and time-dependent input $(S_{in})_0$, that result in growth in the domain of $S$ and $X$ over a large number of time steps.  Figures~\ref{fig:many-X} and ~\ref{fig:many-S} show the substrate and reactant concentrations, respectively, for fifty samples from the training corpus.

\includegraphicsfigure{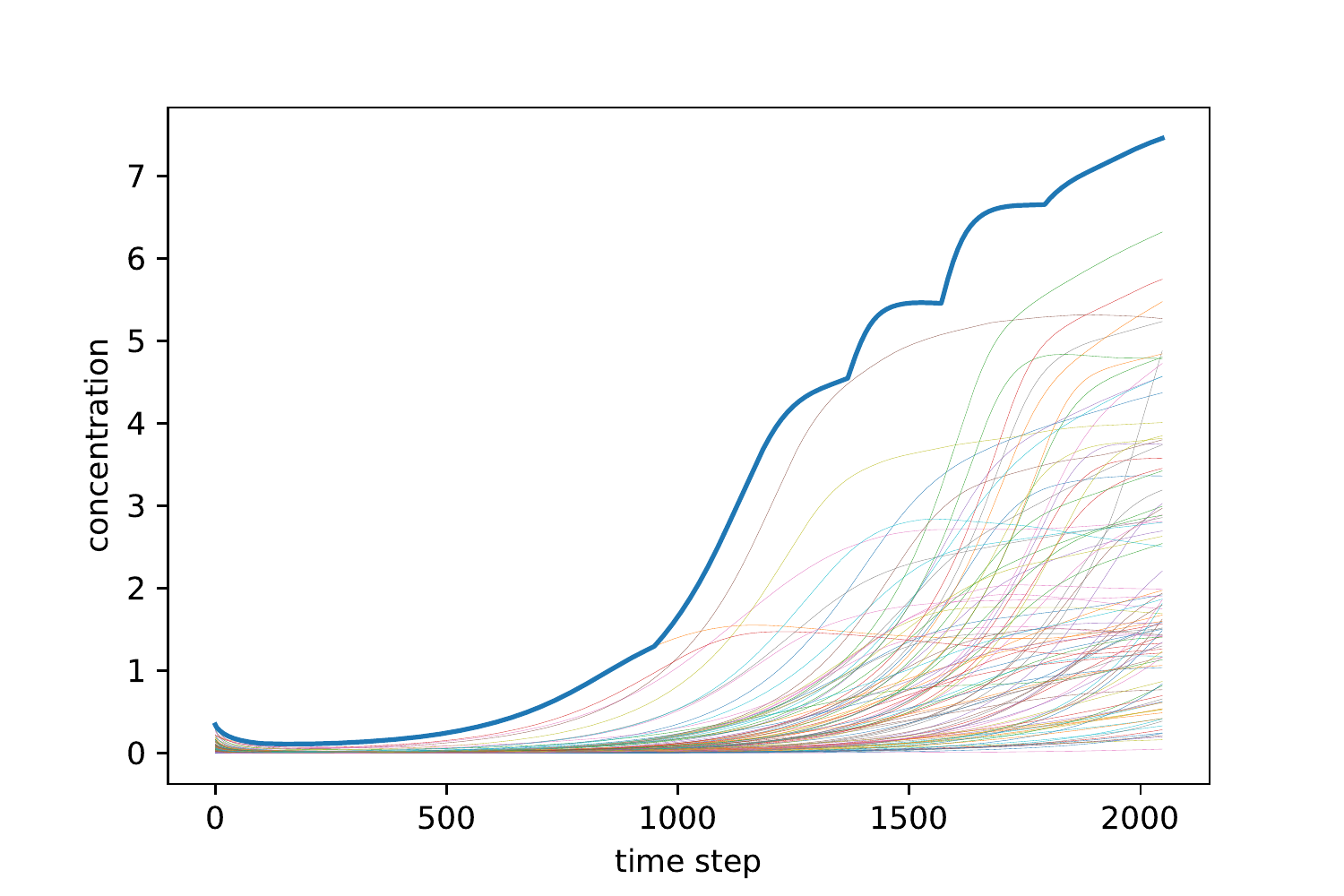}{Concentrations of reactant $X$ for several test-set samples. The topmost curve is the maximum over all samples.}{width=4in}
\includegraphicsfigure{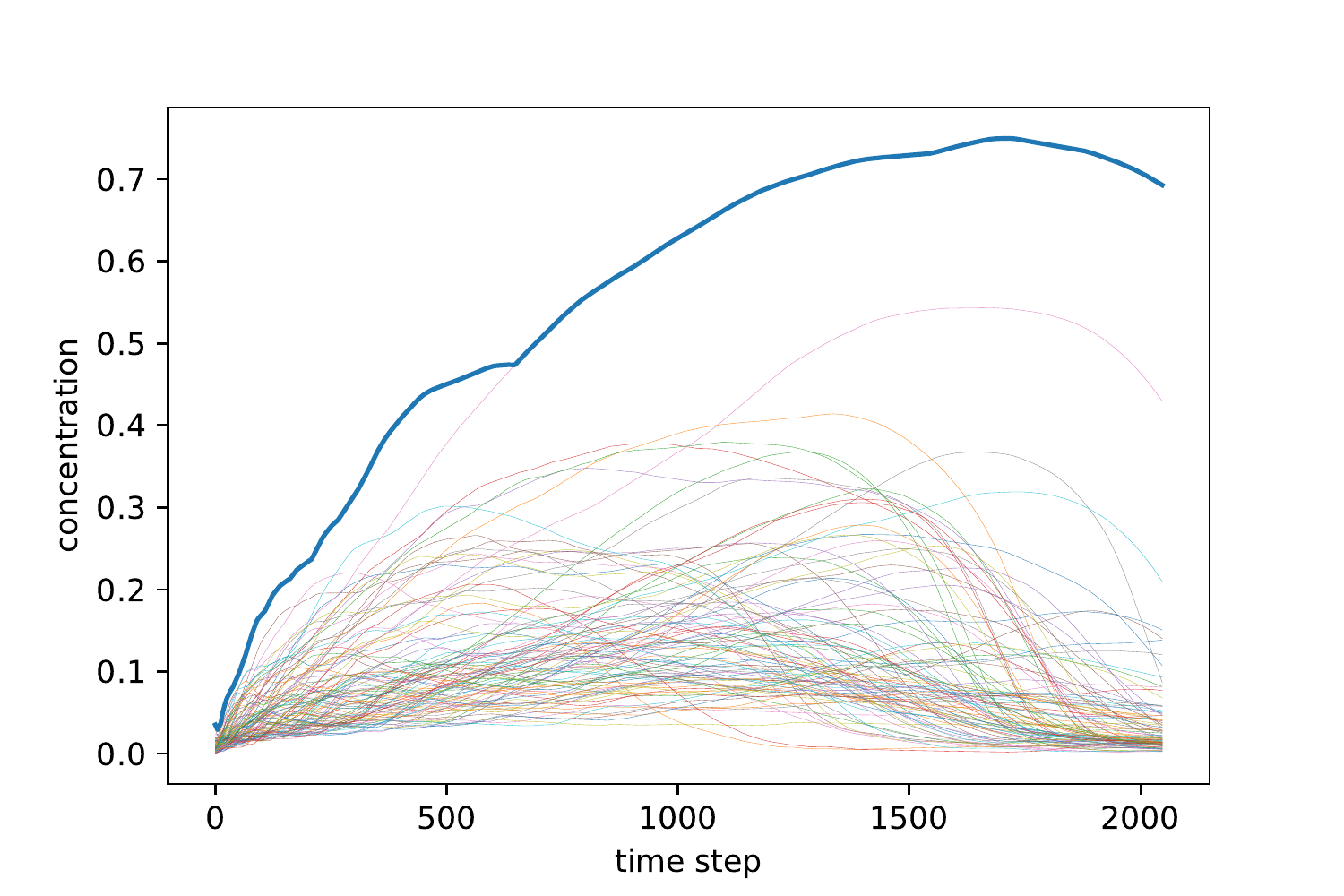}{Concentrations of substrate $S$ for several test-set samples. The topmost curve is the maximum over all samples.}{width=4in}

\section{Results}\label{sec:results}
\subsection{Methodology}


Our networks were trained using TensorFlow's Adam optimization \cite{kingma-ba} implementation. Adam is a momentum-based stochastic optimization method which often achieves good performance for recurrent neural networks. Though Adam has adaptive per-weight learning parameters, it can still be helpful to decay its \lstinline|learning_rate| parameter (which is technically a momentum parameter) as training progresses. While we found that hand-tuning the decay rate of this parameter modestly improved performance in our application, it also introduced a danger of overflow error during training. For the results in our paper we used a fixed learning rate of $10^{-4}$ in order to make meaningful comparisons and avoid overflow in batched training runs.

When evaluating performance after training, the loss function for our network is the sum of the $l_2$-norm losses for the predicted reactant ($X$) and substrate ($S$) concentrations as compared to the ground truth values. During training, only $S$ was used. When directly comparing the reaction rate $\mu$ to its neural network approximation, we also used the $l_2$-norm loss. In order to compare between long and short time-series, target losses are specified per-sample and per-time-step. The ground-truth systems we consider eventually reach a steady state; as the number of time steps becomes large the per-time-step loss average approaches the asymptotic loss. In this regard the average loss can be thought of as somewhat indicative of the asymptotic behavior of the system.

In this work, the following conditions (see \cite{prechelt2012} for a discussion) were monitored as stopping conditions for training.

\begin{itemize}
\item {\em improvement failure:} Twelve epochs pass without improvement in the loss function,
\item {\em generalization failure:} The ratio of the best-so-far loss in the validation set to the current loss in the training set, or {\em generalization error}, exceeds $2.0$,
\item {\em adequate performance:} The average loss per sample per time-step, over all samples in the epoch, is less than $3 \times 10^{-5}$.
\end{itemize}

For simplicity, loss measures in this paper (and in our code) are expressed in terms of the {\em loss ratio}, which in all cases we define as the ratio of the actual loss to adequate performance.

The training, validation, and test sets each contained 1024 entries, each of which consisted of a starting-state triple $(X_0,S_0,V_0)$, and a time-varying-input vector $S_{in}$ with 2048 entries. Initial state values were chosen from normal distributions centered at zero with variances $(.1, .01, 2)$; values for $S_n$ were chosen by choosing $(S_{in})_0$ from a distribution and computing $(S_{in})_k$ by adding a normally distributed value with mean zero and variance $.01$ at each time step. The distributions were chosen to exercise interesting behavior during training, without regard for physical plausibility. We tried to ensure that novel values of the reactant concentration $X$ and the substrate concentration $S$ did not all occur within the first few hundred time steps. This has consequences for training: at the final time steps the most extreme-valued time series lie in regions of the domain of $\mu$ which is relatively poorly trained; and can be a significant source of loss.

We used a feed-forward two-layer linear neural network with ReLU activations (ReLU = restricted linear unit, see \cite{nair-hinton}, \cite{glorot-bordes-bengio}) as our approximator for $\mu$. The ReLU activation function is as follows:
\begin{equation}
\operatorname{relu}(x) = \max(x, 0).
\end{equation}
A linear network with ReLU activations results in a piecewise linear activation. Neural networks with ReLU activations are known to be relatively insusceptible to issues with exploding and vanishing gradients. The implementation of our network is shown in Figure~\ref{fig:feed-forward-code}.

\includecodefigure{feed-forward-code}{Implementation of a two layer feed-forward ReLU network.}

\subsection{Long-time runs}

A two-layer feed-forward network model using ReLU activations was selected to represent the unknown function and trained using time series training data. It follows from \cite{hornik} that with enough nodes in the hidden layer, any function on a compact domain can be well-approximated by a network of this type. Strictly speaking, the result in \cite{hornik} only applies to bounded activation functions. ReLU activations, however, are Lipschitz functions, and one can apply the theorem after noting that $f_t(x) = ReLU(x+t) - ReLU(x)$ is bounded.

The network was trained in two stages. First, in order to produce a reasonable approximation of $\mu$, the input data was coarsened to produce time series of length 256, spanning the same temporal interval. The network was trained on this input until termination (which was by improvement failure at $\tilde 3$ times adequate performance). At this point, the original time series were used to train $\mu$ until improvement failure occurred (approximately at adequate performance). As in later experiments in \cite{psichogios-ungar}, at both stages of computation, loss was computed only for $S$. Effectively, this simulated the situation in which $V$ is unmeasurable.



Ground truth values for $X$ and $S$ for two of the test samples are shown in Figure~\ref{fig:longtime-sample-median-trained} and Figure~\ref{fig:longtime-sample-highest-trained}, along with the the time-series computed for trained $\mu$ using several different time-series lengths during training. The samples chosen were those with the median and highest loss respectively as computed on the time-untruncated test set. The lowest-loss curve is omitted; though the trained curve is barely distinguishable from its corresponding ground truth, the reason is that $X$ stays very close to zero and the curve is not particularly useful as evidence of learning in the presence of hidden data.

Were the time-varying input substrate concentrations $(S_{in})(t)$ constant, the general behavior would be for the reactant to reach a concentration at which it is capable of consuming all incoming substrate, at which point the substrate concentration would drop to $0$ and the system would reach a stable state. As $S_{in}$ was chosen as a random walk, as described above, the region of $X$ for which $\mu$ must train tends to expand as the time series length increases.
  
\includegraphicsfigure{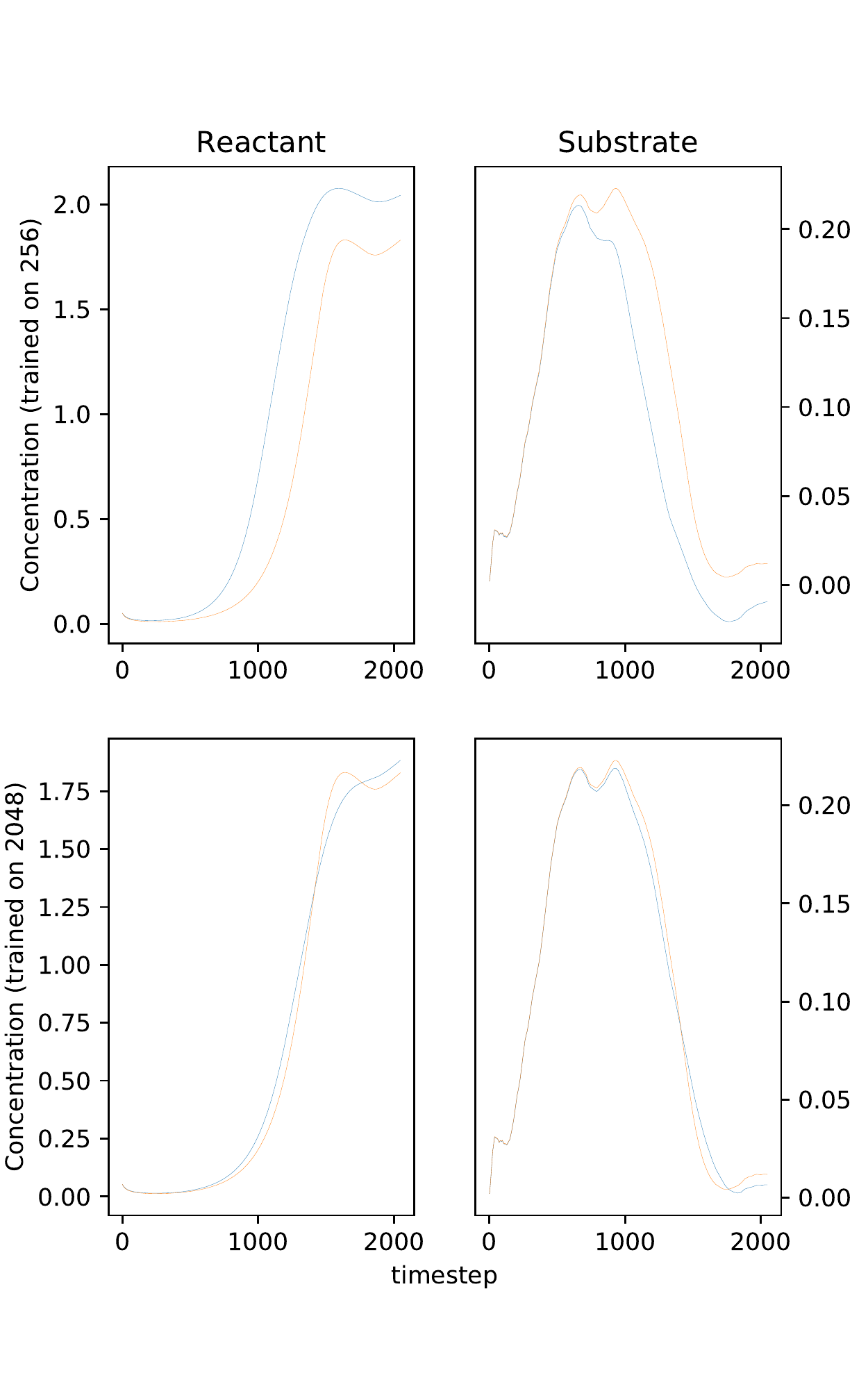}{Median-loss time-series reactant ($X$) and substrate ($S$) concentrations for pretraining (256 time steps) and the full sequence (2048 time steps), with loss function computed from substrate concentration only. The orange curve is the ground truth.}{width=5in}
\includegraphicsfigure{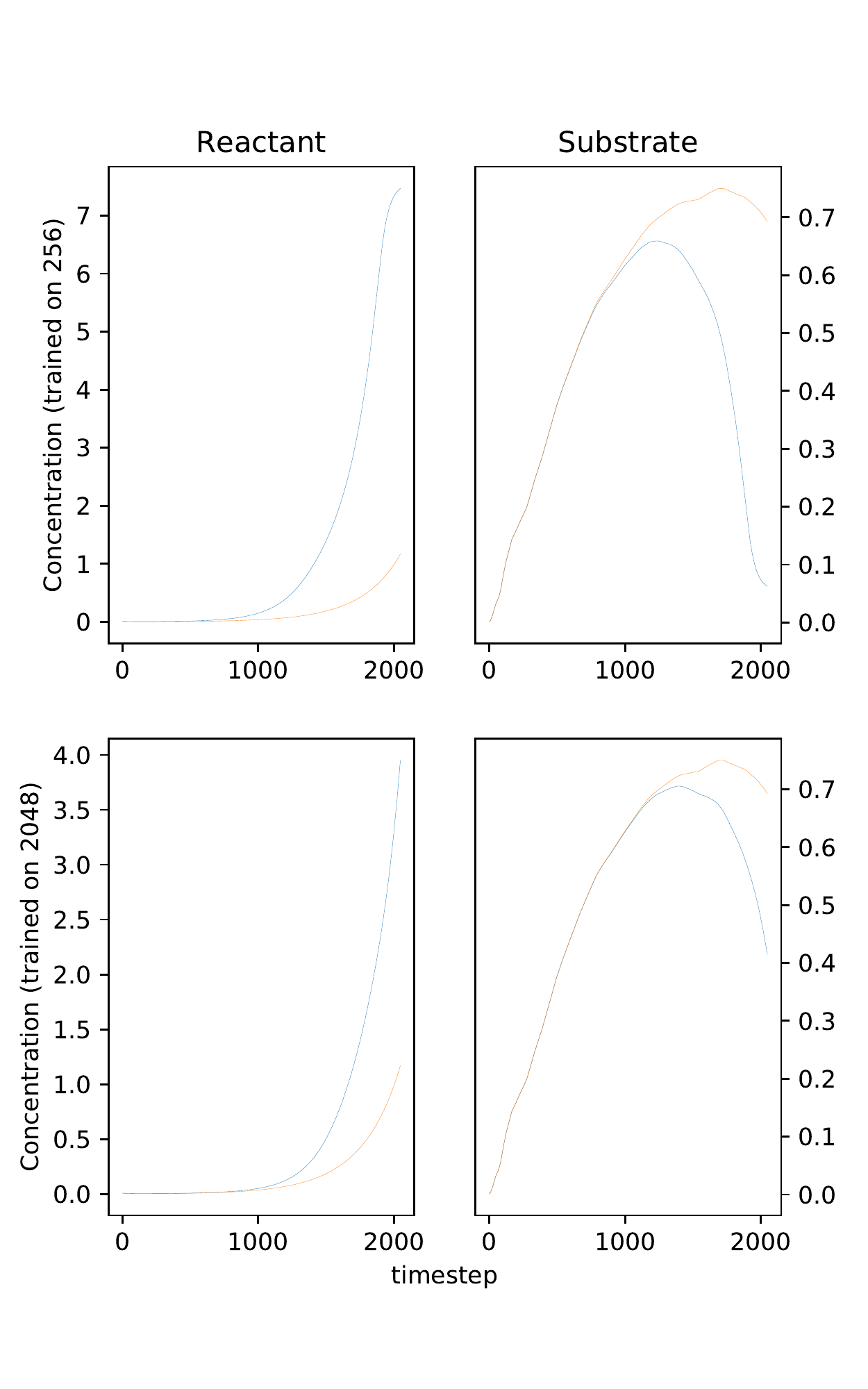}{Highest-loss time-series reactant ($X$) and substrate ($S$) concentrations for pretraining (256 time steps) and the full sequence (2048 time steps), with loss function computed from substrate concentration only. The orange curve is the ground truth.}{width=5in}

\subsection{Missing timestep output}
Next, the loss function was altered so that only every eighth time step (i.e. the 7th, the 15th, etc.) was counted in the loss function. Both $X$ and $S$ were factored into the loss. This simulates a situation in which one has complete state measurements but they do not occur on a small enough time scale to accurately capture the behavior. The network was trained using the same two-stage procedure as before. The network had a bit more difficulty this time; in the first stage terminated with improvement failure at about thirty-eight-times-adequate (squared) loss performance. This translates, at the finer time scale, to roughly five hundred times adequate loss; the network reaches improvement failure in the second stage at about twenty times adequate loss. Timeseries for median and worst-loss solutions are shown in Figure~\ref{fig:omitted-sample-median-trained} and Figure~\ref{fig:omitted-sample-highest-trained}. The lowest-loss curve was similar to that of the previous experiment and is again omitted.

While performance is certainly improvable it should be kept in mind that these executions are a proof of concept intended to demonstrate the capacity of iterated difference equations to learn from long-term time series. Performance is hampered somewhat by lack of a rigorous pretraining scheme. Though it would be desirable to improve fidelity for applications, we take these results as evidence that iterated difference equation networks are capable of learning unknown functions from long time series with missing elements, though the issues we have mentioned (and perhaps others) need to be addressed if high performance is to be achieved. Working out these issues is the subject of ongoing research.

\includegraphicsfigure{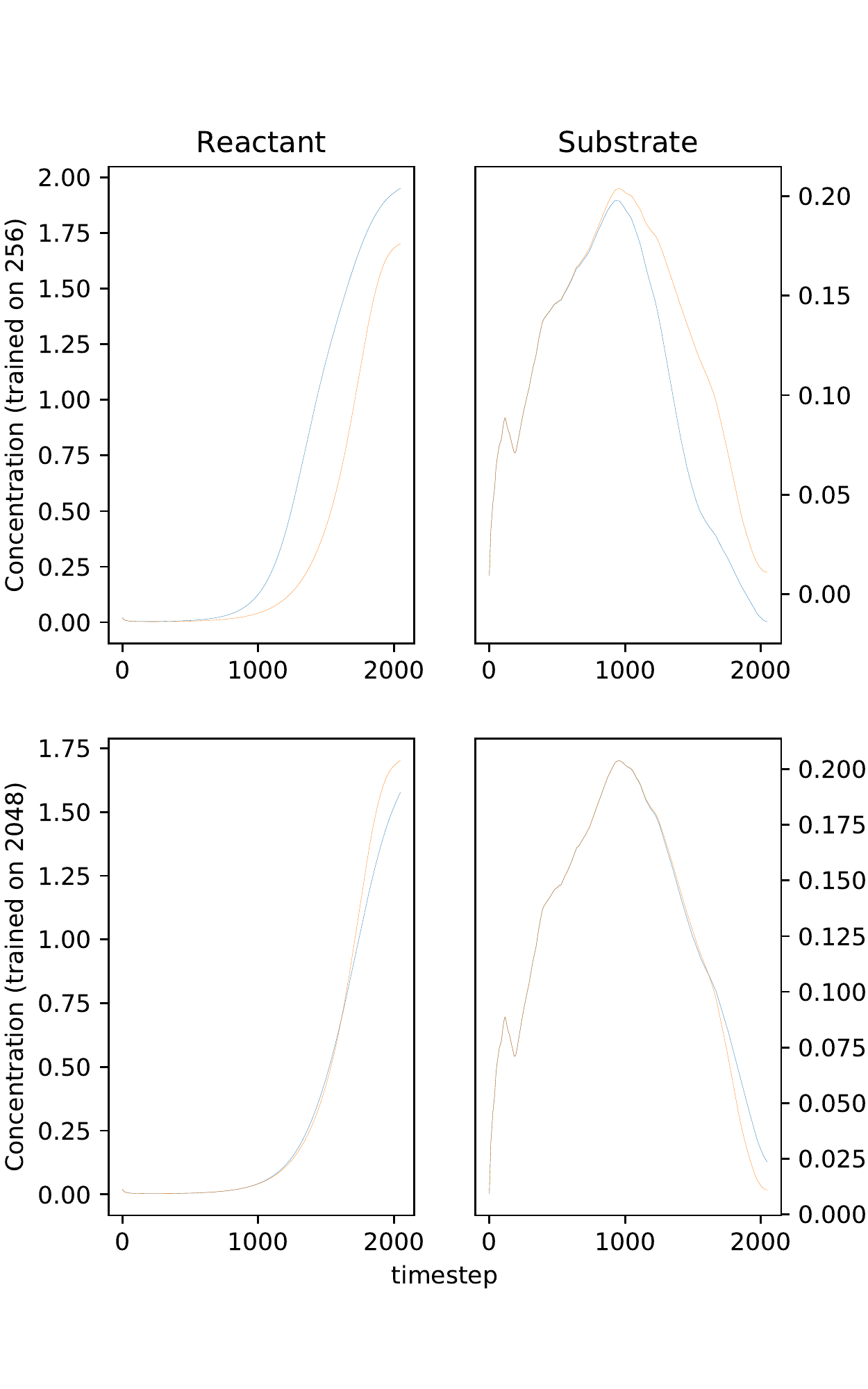}{Median-loss time-series reactant ($X$) and substrate ($S$) concentrations for pretraining (256 time steps) and the full sequence (2048 time steps), with loss function computed using every eighth time step. The orange curve is the ground truth.}{width=5in}
\includegraphicsfigure{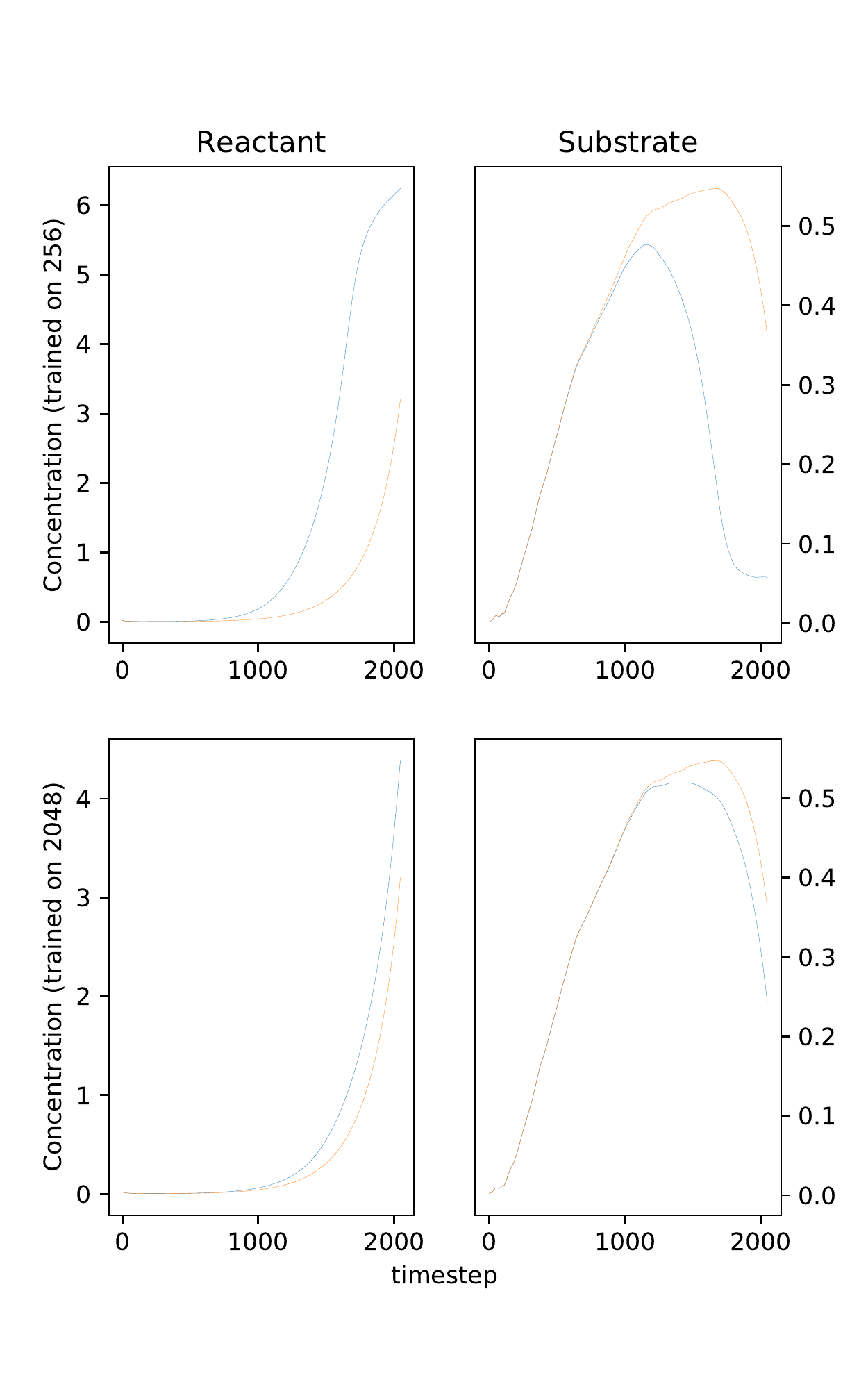}{Highest-loss time-series reactant ($X$) and substrate ($S$) concentrations for pretraining (256 time steps) and the full sequence (2048 time steps), with loss function computed using every eighth time step. The orange curve is the ground truth.}{width=5in}

\section{Comparison of iterated difference equations with other recurrent architectures}\label{sec:comparison}
Though our implementation leverages formal similarities between recurrent networks and iterated difference equations, there are essential differences, as well as in-practice ones. We mention several here.

One important difference concerns the exploding and vanishing gradients problem \cite{hochreiter-bengio-frasconi-schmidhuber}. Exploding and vanishing gradients have been shown to be the cause of slow training for deep networks which must learn long time dependencies. Long time dependencies are learned relationships between temporally separated inputs; the exploding and vanishing gradients problem results from the difficulty of backpropagating an error signal through many time steps. However, the analysis of \cite{hochreiter-bengio-frasconi-schmidhuber} relies on an assumption that does not really hold in our case, namely that long-term dependencies are not accompanied by short-term dependencies which train the same function behavior.  In traditional recurrent network machine learning applications, the network is designed to hold state which is needed for processing inputs several time steps distant. In a differential equation, by contrast, correct behavior over short time scales is equivalent to correct behavior in general.

Another advantage enjoyed by iterated difference equations is that they are often fairly well constrained. The state space is prescribed by the physics of the model and has complexity typically lower than that of the unknown function. The dynamics are also constrained by a known functional form. Additionally, solutions which are not substantially correct will often exhibit incorrect behavior over a large number of time steps, resulting in large error. A recurrent network, in contrast, must often be large enough to train the recurrent layer to serve as a memory which can recognize temporally distant coincidences, and often the distance between events is not known in advance. Good solutions typically require redundancy, are highly non-unique, and can miss essential structure while still offering fairly good performance.

Recurrent networks are deep networks, despite only containing a single layer, because the layer must self-interact over many time-steps in order to produce a correct result. Iterated difference equations are shallow in the sense that the iteration of the recurrent layer can in no way increase the representational capacity of the system. They are also shallow in the sense that any tructation of the layer structure at any time step produces a sensible result. It is often times useful to think of iterated difference equation networks as shallow networks that are trained using deep techniques.

For unitary physical systems, techniques and analysis from the unitary recurrent neural network literature (e.g. \cite{wisdom-powers-hershey-leroux-atlas}) are naturally relevant as the unitarity is a physically motivated constraint. However, we did not pursue this direction in the current work.

As with other recurrent networks\cite{glorot-bengio}, our iterated difference equations are subject to difficulties with initialization. In our case these issues are driven by the functional form of the differential equation rather than the exploding and vanishing gradients problem. The ReLU network outlined above models a piecewise linear function. Considering the fedbatch reactor model (\ref{eq_X}) and (\ref{eq_S}). One can easily see that if $\mu$ is unfortunately chosen, for example if $\mu$ takes positive values on the second quadrant of the $X-S$ plane, increasing as one moves away from the origin, then $X$ may increase while $S$ decreases in a manner that is self-reinforcing, resulting in super-exponential blowup. Generally, one must either make fortunate choices, constrain the trained function $\mu$ so that exponential dynamics are avoided, or pretrain $\mu$ so that it is a close enough approximation to (presumably well-behaved) $\mu$ that it is itself well-behaved. Due to the exploratory nature of this paper we chose to rely on the first method.

There are other issues which can affect initialization. It may be necessary for $\mu$ to be accurate over multiple time scales, in which case a uniform (or normal) distribution of initial bias values may be less suitable than another choice such as a log-uniform or log-normal distribution. In any case, if the input to $\mu$ is multidimensional, the different inputs may have different scales and scaling laws.

\section{Conclusion}

We have constructed a scalable system for approximating unknown functions in iterated differential equations. This system avoids solving the sensitivity equations, and thus, integration with a differential equation solver package that must be invoked at each time-step during back propagation. Our networks show the ability to reconstruct unknown functions when data is categorically excluded or time-intermittent.

Our networks are formally quite similar to recurrent neural networks and suffers from some of the same issues. However, from the perspective of ease of training, our networks have many advantages, due to their constrained dynamics, the (often) uniqueness of correct solutions, and the ability of partial timeseries or aggregated time series to be interpreted sensibly and used as training data. It would be interesting to develop rubust initialization methods in this context; it may well be that the specific formal properties of our systems allow more to be said than in the recurrent network case. Finally, since our networks are relatively simple in structure they offer simple and functional examples of some relatively deep recurrent neural networks.

\bibliographystyle{plain}
\bibliography{references.bib,pde.bib}

\end{document}